\documentclass{article}

\usepackage{microtype}
\usepackage{graphicx}
\usepackage{subfigure}
\usepackage{multirow}
\usepackage{amssymb}

\usepackage{booktabs} 

\usepackage{hyperref}

\usepackage[accepted]{icml2019}

\icmltitlerunning{SelectiveNet: A Deep Neural Network with an Integrated Reject Option}

\begin{document}

\twocolumn[
\icmltitle{SelectiveNet: A Deep Neural Network with an Integrated Reject Option}



\icmlsetsymbol{equal}{*}

\begin{icmlauthorlist}
\icmlauthor{Yonatan Geifman}{tech}
\icmlauthor{Ran El-Yaniv}{tech}
\end{icmlauthorlist}

\icmlaffiliation{tech}{Technion - Israel Institute of Technology}

\icmlcorrespondingauthor{Yonatan Geifman}{yonatan.g@cs.technion.ac.il}

\icmlkeywords{Machine Learning, ICML}

\vskip 0.3in
]



\printAffiliationsAndNotice{}  

\begin{abstract}
We consider the problem of \emph{selective prediction} (also known as \emph{reject option}) in deep neural networks, and introduce SelectiveNet, a deep neural architecture with an integrated reject option. Existing rejection mechanisms are  based mostly on a threshold over the prediction confidence of a pre-trained network. In contrast, SelectiveNet is trained to optimize both classification (or regression) and rejection simultaneously, end-to-end.  The result is a deep neural network that is optimized over the covered domain. In our experiments, we show a consistently improved risk--coverage trade-off over several well-known classification and regression datasets, thus reaching  new state-of-the-art results for deep selective classification.

\end{abstract}


\newtheorem{theorem}{Theorem}[section]
\newtheorem{lemma}[theorem]{Lemma}
\newtheorem{corollary}[theorem]{Corollary}
\newtheorem{observation}[theorem]{Observation}

\newtheorem{assumption}[theorem]{Assumption}
\newtheorem{example}[theorem]{Example}

\newenvironment{proofsketch}[1][Proof Sketch:]{\begin{trivlist} \item[\hskip \labelsep
		{\bfseries #1}]}{\hfill{$\Box$}\end{trivlist}}

\newcommand{\argmax}{\operatornamewithlimits{argmax}}
\newcommand{\argmin}{\operatornamewithlimits{argmin}}

\newcommand{\rnote}[1]{ {\color{red}  [ Ran: \textit{#1}]} }
\newcommand{\ynote}[1]{ {\color{blue} [ Yonatan: \textit{#1}]} }
\newcommand{\sharon}[1]{ {\color{blue} {#1}} }

\newcommand{\vect}[1]{\mathbf{#1}}
\newcommand{\mb}[1]{{\mathbf{\red{#1}}}}
\newcommand{\Bf}[1]{{\mathbf{#1}}}

\newcommand{\one}{\mathbf{1}}

\newcommand{\hr}{\hat{r}}
\newcommand{\hphi}{\hat{\phi}}

\newcommand{\VS}{\mathcal{VS}} 

\newcommand{\data}[1]{\texttt{#1}}
\newcommand{\tdata}[1]{\scriptsize{\texttt{#1}}}
\newcommand{\COIL}{\data{COIL}}
\newcommand{\MUSH}{\data{MUSH}}
\newcommand{\MUSK}{\data{MUSK}}
\newcommand{\PIMA}{\data{PIMA}}
\newcommand{\BUPA}{\data{BUPA}}
\newcommand{\VOTING}{\data{VOTING}}
\newcommand{\MONK}{\data{MONK}}
\newcommand{\IONO}{\data{IONOSPHERE}}
\newcommand{\TAE}{\data{TAE}}
\newcommand{\DIGIT}{\data{DIGIT}}
\newcommand{\TEXT}{\data{TEXT}}

\newcommand{\sCOIL}{\tdata{COIL}}
\newcommand{\sMUSH}{\tdata{MUSH}}
\newcommand{\sMUSK}{\tdata{MUSK}}
\newcommand{\sPIMA}{\tdata{PIMA}}
\newcommand{\sBUPA}{\tdata{BUPA}}
\newcommand{\sVOTING}{\tdata{VOTING}}
\newcommand{\sMONK}{\tdata{MONK}}
\newcommand{\sIONO}{\tdata{IONOSPHERE}}
\newcommand{\sTAE}{\tdata{TAE}}
\newcommand{\sDIGIT}{\tdata{DIGIT}}
\newcommand{\sTEXT}{\tdata{TEXT}}

\newcommand{\myalg}[1]{\texttt{#1}}
\newcommand{\mytalg}[1]{\scriptsize \texttt{#1}}
\newcommand{\SGT}{\myalg{SGT}}
\newcommand{\ZHU}{\myalg{GRMF}}
\newcommand{\BELKIN}{\myalg{SOFT}}
\newcommand{\SCHOLKOPF}{\myalg{CM}}
\newcommand{\EXPP}{\myalg{+EXPLORE}}
\newcommand{\sEXPP}{\mytalg{+EXPLORE}}
\newcommand{\sZHU}{\mytalg{STRICT}}
\newcommand{\sBELKIN}{\mytalg{SOFT}}
\newcommand{\sSCHOLKOPF}{\mytalg{RANGE}}
\newcommand{\sSGT}{\mytalg{SGT}}

\newcommand{\ENG}{\myalg{ENERGY}} 

\newcommand{\KNN}{\myalg{kNN}}

\newcommand{\comp}[1]{\small \texttt{#1}}
\newcommand{\QEXPLORE}{\comp{Q-EXPLORE}}
\newcommand{\QEXPLOIT}{\comp{Q-EXPLOIT}}
\newcommand{\EXPPSWITCH}{\comp{EXPLORE-EXPLOIT-SWITCH}}

\newcommand{\cA}{{\cal A}}
\newcommand{\cB}{{\cal B}}
\newcommand{\cF}{{\cal F}}
\newcommand{\cG}{{\cal G}}
\newcommand{\cH}{{\cal H}}
\newcommand{\cL}{{\cal L}}
\newcommand{\cV}{{\cal V}}
\newcommand{\cX}{{\cal X}}
\newcommand{\hL}{{\widehat{L}}}
\newcommand{\hcL}{{\widehat{\mathcal{L}}}}
\newcommand{\hR}{\widehat{R}}
\newcommand{\bB}{\mathbf{B}}
\newcommand{\bW}{\mathbf{W}}
\newcommand{\bR}{\mathbf{R}}
\newcommand{\bD}{\mathbf{D}}
\newcommand{\bG}{\mathbf{G}}
\newcommand{\bL}{\mathbf{L}}
\newcommand{\bI}{\mathbf{I}}
\newcommand{\bC}{\mathbf{C}}
\newcommand{\bZ}{\mathbf{Z}}
\newcommand{\ba}{\mathbf{a}}
\newcommand{\bd}{\mathbf{d}}
\newcommand{\be}{\mathbf{e}}
\newcommand{\bg}{\mathbf{g}}
\newcommand{\Bg}{\mathbf{g}}
\newcommand{\bh}{\mathbf{h}}
\newcommand{\bp}{\mathbf{p}}
\newcommand{\bq}{\mathbf{q}}
\newcommand{\bt}{\mathbf{t}}
\newcommand{\bu}{\mathbf{u}}
\newcommand{\bv}{\mathbf{v}}
\newcommand{\bx}{\mathbf{x}}
\newcommand{\by}{\mathbf{y}}
\newcommand{\bz}{\mathbf{z}}
\newcommand{\E}{\mathbf{E}}
\newcommand{\var}{\text{var}}
\renewcommand{\H}{\mathbf{H}}
\renewcommand{\S}{\mathbf{S}}
\newcommand{\T}{\mathbf{T}}
\newcommand{\CM}{\mathtt{CM}}
\newcommand{\CMRAD}{\mathtt{CM-SUP}}

\newcommand{\nchoosek}[2]{\left(\begin{array}{c}#1\\#2\end{array}\right)}
\renewcommand{\Pr}{\mathbf{Pr}}
\newcommand{\head}{\mathrm{head}}
\newcommand{\tail}{\mathrm{tail}}
\newcommand{\rad}{\mathtt{R}}
\newcommand{\parr}{\mathtt{Par}}
\newcommand{\tA}{\mathtt{A}}
\newcommand{\bpi}{\pmb{\pi}}
\newcommand{\cN}{{\mathcal N}}
\newcommand{\cY}{{\mathcal Y}}
\newcommand{\rE}{\mathbf{E}}
\newcommand{\pig}{\tilde\pi_{\gamma/k}}
\newcommand{\al}{\alpha}
\newcommand{\QED}{\hfill{$\Box$}}
\newcommand{\lam}{\lambda}
\newcommand{\red}[1]{\textcolor{red}{#1}}
\newcommand{\err}{\mathop{\rm er}}

\newcommand{\I}{\mathbb{I}}
\newcommand{\hf}{f_Q}
\newcommand{\hg}{\hat{g}}
\newcommand{\LA}{{\mathcal L }}
\newcommand{\GLi}{G_L^{(i)}}
\newcommand{\GUi}{G_U^{(i)}}
\newcommand{\fail}{\texttt{fail}}
\newcommand{\reals}{\mathbb{R}}
\newcommand{\eqdef}{\triangleq}


\section{Introduction}
\label{introduction}

Detecting and controlling statistical uncertainties in machine learning processes is essential in many mission-critical machine learning applications such as autonomous driving, medical diagnosis or home robotics.
In \emph{selective prediction} our goal is to learn predictive models that know what they do not know. These models are allowed to abstain whenever they are not sufficiently 
confident in their prediction. 

The idea of a reject option was already studied over 60 years ago by \citet{chow1957optimum}.
  Reject option mechanisms have been considered for many hypothesis classes and 
learning algorithms. These mechanisms are mainly driven by two related ideas. The first is 
the cost-based model whereby the selective model is optimized to minimize a loss function that tracks a user-specified cost for abstention. Other mechanisms are
built upon  functions that serve as proxies for prediction confidence. Typical examples of these proxies are functional margin, distance to nearest neighbor(s), etc.

In this work we focus on selective models for deep neural networks (DNNs).
The most  recent relevant paper on this topic is \citet{geifman2017selective}, which 
shows how to construct a probabilistically-calibrated selective classifier 
using any certainty estimation function for a given (already trained)
model. Interestingly, among the best certainty proxies for a trained 
network is the simple Softmax Response (SR) estimator, which is often used
by practitioners. 

Within an abstention-constrained framework, whereby the user specifies the desired 
target coverage, we optimize, end-to-end, a neural network with an integrated 
reject option, designed to be optimal for the required coverage slice.
Our technique yields effective selective models that consistently improve
all known techniques. 

From a learning-theoretic perspective, it is not hard to argue that learning a specialized rejection model for each coverage slice is potentially better than rejecting based on confidence rates extracted from a pre-trained network.
To demonstrate the advantage of such specialized training, consider the following vignette. Bob and Alice are about to take an exam at the end of their machine learning course. It is announced that the exam will contains 10 questions, one for each of the course topics, but the students will be asked to choose exactly five questions. Alice decides to cover only the easiest five topics while Bob's mom forces him to study all ten topics. Who do you think is going to get the higher grade?

The contributions of this paper are:
\begin{itemize}
\item A selective loss function that optimizes a specified coverage slice using a variant of the interior point optimization method \cite{potra2000interior}.
\item SelectiveNet: A three-headed network for end-to-end learning of selective 
classification models.
\item SelectiveNet for regression: Providing the first alternative to costly methods such as MC-dropout or ensemble techniques.
\item An empirical advantage demonstrated over four datasets for classification and regression with a significant advantage over MC-dropout 
and SR.
\end{itemize}

\section{Selective Prediction Problem Formulation}
\label{sec:problem setting}

In this work we consider selective prediction problems of any type, classification or regression. A supervised prediction task is formulated as follows.
Let $\cX$ be any feature space and $\cY$ a label space. For example,
in classification, 
$\cX$ could be images, and $\cY$ could be class labels in the case of classification
or the coordinates of a bounding box in the case of regression.
Let $P(X,Y)$ be a distribution over $\cX \times \cY$.
A model,  $f : \cX \to \cY$,  is called a \emph{prediction function}, and its \emph{true risk}  w.r.t. $P$ is
$R(f) \eqdef E_{P(X,Y)}[\ell(f(x),y)]$, where 
$\ell : Y \times Y \to \reals^+$ is a given loss function, for example, 
the 0/1 error or the squared loss.
Given a labeled set
$S_{m}=\{(x_{i},y_{i})\}_{i=1}^{m}\subseteq(\cX\times \cY)^m$ sampled i.i.d. from $P(X,Y)$,
the \emph{empirical risk} of the classifier $f$ is
$\hr(f | S_m) \eqdef \frac{1}{m}\sum_{i=1}^{m}\ell(f(x_{i}),y_{i})$.

A \emph{selective model} \cite{ElYaniv10} is a pair $(f,g)$, where $f$ is a \emph{prediction function},  and $g: \cX\rightarrow\{0,1\}$ is a \emph{selection function}, which serves as a binary qualifier 
for $f$ as follows,

\begin{eqnarray*}
\label{eq:selectiveClassifier}
(f,g)(x) \eqdef \left\{
             \begin{array}{ll}
             f(x), & \hbox{if $g(x)=1$;} \\
               \textnormal{don't know}, & \hbox{if $g(x)=0$.} 
             \end{array}
           \right.
\end{eqnarray*}

Thus, the selective model abstains from prediction at a point $x$ iff $g(x) = 0$.
A \emph{soft} selection function can also be considered, where  $g: \cX\rightarrow[0,1]$, and decisions can be taken probabilistically or deterministically (w.r.t. a threshold).
The performance of a selective model is quantified using \emph{coverage} and \emph{risk}.
Coverage is defined to be 
$$\phi(g) \eqdef E_{P}[g(x)],$$
the probability mass of the non-rejected region in $\cX$.
The selective risk of $(f,g)$ is  
$$
R(f,g) \eqdef \frac{E_{P}[\ell(f(x),y)g(x)]}{\phi(g)}.
$$
Clearly, the risk of a selective model can be traded-off for coverage. The entire performance profile 
of such a model can be specified by its \emph{risk--coverage} curve, defined to be the risk as a function of coverage 
\cite{ElYaniv10, wiener2012pointwise}.

The true selective risk and  true coverage have corresponding empirical counterparts that can be calculated for any given labeled set $S_m$.  The \emph{empirical selective risk} is 
$$
\hr(f,g | S_m) \eqdef \frac{\frac{1}{m}\sum_{i=1}^{m}\ell(f(x_{i}),y_{i}) g(x_i)}{\hat{\phi}(g | S_m)},
$$
and the \emph{empirical coverage} is
$$\hphi(g | S_m) \eqdef \frac{1}{m}\sum_{i=1}^{m}g(x_i).$$

An optimal selective model could be defined in one of two ways:  by either optimizing the selective risk, given a constraint on the coverage or vice versa. In this work, we focus on the first case, which is defined as follows.
For a given coverage rate $0<c\leq 1 $ and hypothesis class $\Theta$, the optimal selective model is 
\begin{eqnarray}
\label{eq:objective}
\theta^* = \arg\min_{\theta\in \Theta}(R(f_\theta,g_\theta))
\\
s.t.\textnormal{  } \phi(g_\theta)\geq c \nonumber.
\end{eqnarray}

In this paper, $\Theta$ is a set of parameters for a given deep network architecture for $f$ and $g$. 

It is possible to convert any selective model defined as above (with controlled coverage and optimized risk) to a model that controls risk and optimizes coverage using the technique of \citet{geifman2017selective}.

\section{Related Work}
The literature on reject option techniques is quite extensive and mostly focuses on 
hypothesis classes and learning algorithms such as SVMs, nearest neighbours and boosting \cite{fumera2002support,hellman1970nearest,cortes2016boosting,wiener2015agnostic}. A common approach to build selective predictors is 
to construct and add a selection mechanism to a trained prediction model. 
Another approach, first proposed by
\citet{cortes2016learning}, is to jointly learn the
predictor and selection function. In this paper
we follow this approach.
When considering 
selective prediction in neural networks (NNs), a straightforward and 
effective technique is to set a threshold over a confidence score derived from 
a pre-trained network, which was already optimized to predict all points \cite{cordella1995method,de2000reject,wiener2012pointwise}. 
\citet{geifman2017selective}, extending this technique for deep neural networks (DNNs), show how to derive a selective classifier based on any confidence score function. They also show how to compute a high confidence guarantee on the selective risk at test time.
In the latter work, two techniques for extracting 
confidence scores from an NN are discussed. The first is the Softmax Response (SR), 
defined to be the maximal activation in the softmax layer (in classification architectures). The second confidence score is the Monte-Carlo dropout (MC-dropout)
technique of \citet{gal2016dropout}. MC-dropout estimates prediction confidence based on the statistics of numerous feed-forward passes through the network with dropout applied.
To the best of our knowledge, the MC-dropout is the only non-Bayesian technique
to extract a confidence score for a single network. Unfortunately, the prediction cost of an  effective MC-dropout can reach hundreds of feed-forward iterations for each 
prediction.

Another family of confidence score functions that can be used for selective classification is based on statistics of ensembles of multiple models \cite{lakshminarayanan2017simple, geifman2018bias}. In the present work, we focus on selective classification with a single classifier while noting that extending our work for ensembles of selective classifiers is straightforward and likely to improve the results.

\section{SelectiveNet}
\label{sec:selectivenet}
In this section we first define \emph{SelectiveNet}, a deep neural architecture  allowing  end-to-end optimization of selective models.  In conjunction, we also  propose a suitable loss function and optimization procedure to train it for any desired target coverage. 

\subsection{SelectiveNet Architecture}
\begin{figure}[htb]
	\centering
	\includegraphics[width=0.8\linewidth]{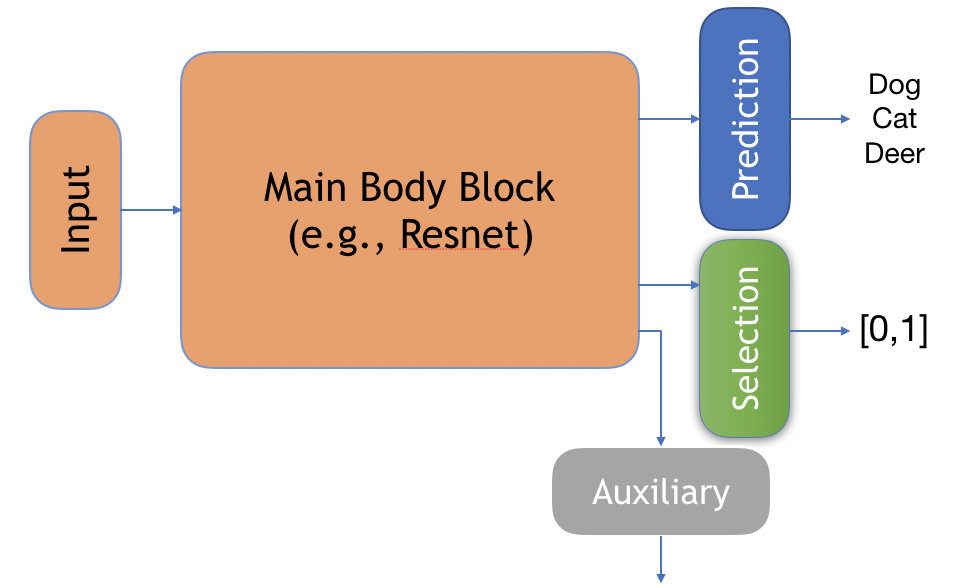}
	\caption{The SelectiveNet schematic architecture}
	\label{fig:architecture}
\end{figure}

SelectiveNet is a selective model $(f,g)$ that optimizes both  $f(x)$ and $g(x)$ in a single DNN model. A schematic view of SelectiveNet is depicted in  Figure~\ref{fig:architecture}.  The input is processed by  the main body block, consisting of a number of DNN layers or sub-blocks. The last layer of the main body is termed the  \emph{representation layer}. The main body block can be assembled using any type of architecture relevant to the problem at hand (e.g., convolutional, fully connected or recurrent architectures).
SelectiveNet has three output heads for prediction, selection and auxiliary prediction.
The role of the selection head is to implement the selection function $g(x)$. 
The role of the prediction head is to implement the prediction $f(x)$, 
and the auxiliary prediction  head (denoted as $h(x)$) predicts a related prediction task 
in order to enrich or enforce the construction of relevant features in the main body.
The auxiliary head is only used for training and its role will be clarified later after we define the loss function.
The architectures of these three heads can vary depending on the task type and complexity but it is always the case that the final layer of the selection head $g(x)$ is a single neuron with a sigmoid activation. The final layer of $f(x)$ depends on the application and could be softmax (classification) or linear (regression).
At inference time, a sample $x$ is fed to SelectiveNet, which predicts $f(x)$ if and only if  $g(x) \geq 0.5$; otherwise, SelectiveNet abstains from predicting the label of $x$.

\subsection{SelectiveNet Optimization}
The selective prediction objective is given in Equation~(\ref{eq:objective}).  To enforce the coverage constraint, we utilize a variant of the well-known Interior Point Method (IPM) \cite{potra2000interior}. This results in the following unconstrained objective, which is averaged over the samples in $S_m$,
\begin{eqnarray}
\label{eq:ourLoss}
\cL_{(f,g)}&\eqdef&\hat{r}_\ell(f,g|S_m) +\lambda \Psi (c-\hat{\phi}(g|S_m))\\
\Psi(a)&\eqdef&\max(0,a)^2, \nonumber
\end{eqnarray}
where $c$ is the target coverage,  $\lambda$ is a hyperparameter controlling the relative importance of the constraint, and $\Psi$ is  a quadratic penalty function. The choice of $\lambda$ is discussed in Section~\ref{sec:implementation}. In this work we train the auxiliary head, $h$, using the same prediction task as assigned to $f$ (classification or regression) using a standard loss function, $$\cL_{h}=\hat{r}(h|S_m)=\frac{1}{m}\sum_{i=1}^{m}\ell(h(x_i),y_i),$$
which is oblivious to any coverage considerations. We optimize SelectiveNet using a convex combination of the selective loss $\cL_{(f,g)}$ (\ref{eq:ourLoss}) and the auxiliary loss $\cL_{h}$. Thus, the overall training objective is 
$$\cL=\alpha \cL_{(f,g)}+(1-\alpha)\cL_h.$$

The use of the auxiliary head, $h$, is essential to optimizing SelectiveNet. Without $h$, SelectiveNet will focus on a fraction $c$ of the training set, before accurate low level features are constructed. In such a case, SelectiveNet will tend to overfit to the wrong subset of the training set.
The auxiliary head exposes the main body block to all training instances throughout the training process. For example, when taking the parameter alpha to be very small, SelectiveNet will result with poor selective risk. In all our experiment we used $\alpha=0.5$ without any hyperparameter optimization.

\begin{table*}[h!]
	\centering
	\begin{tabular}{c c c} 
		\hline
		\hline
		Target coverage &  SR test coverage & SelectiveNet test coverage\\
		\hline\hline
0.70 & 60.19$\pm$0.23 & 77.34$\pm$0.05   \\ [0.5ex]  
0.75 & 64.16$\pm$0.26 & 80.17$\pm$0.11  \\ [0.5ex]  
0.80 & 68.01$\pm$0.36 & 82.63$\pm$0.12   \\ [0.5ex]  
0.85 & 72.05$\pm$0.30 & 85.34$\pm$0.24   \\ [0.5ex]  
0.90 & 76.41$\pm$0.26 & 87.74$\pm$0.32   \\ [0.5ex]  
0.95 & 82.30$\pm$0.23 & 90.99$\pm$0.67   \\ [0.5ex]  
    		\hline\hline
Average violation & 11.98 & 3.63\\
\hline

	\end{tabular}
	\caption{Selective test coverage rates (\%) for various trained coverage rates for SR and SelectiveNet for the Cifar-10 dataset.}
	\label{table:coverage_accuracy}
\end{table*}

\section{Coverage Accuracy}

When considering learning with constrained ERM as in Equation~(\ref{eq:objective}), we expect to see violations of the constraint over a test set. In our case, this means that the true coverage, $\phi(g)$, might be smaller than the desired target coverage $c$. On the other hand, even if the constraint is not violated, we are susceptible to 
sub-optimal models, $(f,g)$, where $\phi(g) > c$ (the optimal selective model must 
satisfy the constraint with equality). When dealing with deep neural models, such  coverage inaccuracies can be amplified due to the high capacity of these models.

Indeed, consider Table~\ref{table:coverage_accuracy} showing the 
resulting test coverage obtained using both SelectiveNet and SR.
Here, SelectiveNet was optimized with target coverage rates 
$c \in \{0.7, 0.75, \ldots, 0.9\}$, and the coverage of SR was controlled by choosing 
an appropriate rejection threshold over the training set.
 Clearly, both SR and SelectiveNet violate the target coverage rate
 in all cases. For example, when the target coverage is 75\%, the realized coverage of SR is 64.16\% and that of SelectiveNet is  80.17\%. 
 
 While the coverage accuracy of SelectiveNet is significantly better than that of SR (average violation 3.625\% vs. 11.98\%, see Table~\ref{table:coverage_accuracy}), we are interested in
 generating selective models whose test coverage is as close as possible to the target 
 coverage. This goal can be  achieved easily
 using the following 
simple post-training coverage calibration technique, which relies on an independent \emph{unlabeled} validation set $V_n$ containing $n$ samples.
To calibrate to the target coverage, we 
estimate an appropriate threshold $\tau$ for $g(x)$ values and predict using the following decision rule
\begin{eqnarray*}
\label{eq:selectiveClassifier}
(f,g_{\tau})(x) \eqdef \left\{
             \begin{array}{ll}
             f(x), & \hbox{if $g(x)\geq \tau$} \\
               \textnormal{don't know}, & \hbox{otherwise.} 
             \end{array}
           \right.
\end{eqnarray*}
Given the validation set $V_n$,  $\tau$ is set to be the 
$100(1-c)$ percentile of the distribution of $g(x_i)$, $x_i \in V_n$. 

Noting that the event $g(x_i) \geq \tau$ is a Bernoulli variable, we can 
straightforwardly use the Hoeffding bound \cite{Hoeffding:1963} to bound the probability of (double sided) coverage violation greater than $\epsilon$, 
$$
\Pr\{\epsilon\textnormal{\rm -violation}\} \leq 2 e^{2n\epsilon^2} .
$$
Equating the right-hand side to a confidence parameter $\delta$, we get that with probability at least $1-\delta$, the resulting selective model will be in the range $[c-\epsilon, c+\epsilon]$ where $\epsilon = \sqrt{\ln(2/\delta)/(2n)}$.

\begin{table*}[h!]
	\centering
	\begin{tabular}{c c c c c c} 
		\hline
		\hline
		Coverage&  SelectiveNet risk & MC-dropout risk&  \% improvement &SR risk & \% improvement\\
		\hline\hline
1.00 & 6.79$\pm$0.03 & 6.79$\pm$0.03 & 0.00  & 6.79$\pm$0.03 & 0.00   \\ [0.5ex]
0.95 &\textbf{ 4.16$\pm$0.09} & 4.58$\pm$0.05 & 8.98  & 4.55$\pm$0.07 & 8.56   \\ [0.5ex]
0.90 & \textbf{2.43$\pm$0.08} & 2.92$\pm$0.01 & 16.99  & 2.89$\pm$0.03 & 16.14   \\ [0.5ex]
0.85 & \textbf{1.43$\pm$0.08} & 1.82$\pm$0.09 & 21.35  & 1.78$\pm$0.09 & 19.79   \\ [0.5ex]
0.80 & \textbf{0.86$\pm$0.06} & 1.08$\pm$0.05 & 20.39  & 1.05$\pm$0.07 & 17.87   \\ [0.5ex]
0.75 & \textbf{0.48$\pm$0.02} & 0.66$\pm$0.05 & 26.86  & 0.63$\pm$0.04 & 22.71   \\ [0.5ex]
0.70 &\textbf{ 0.32$\pm$0.01} & 0.43$\pm$0.05 & 26.38  & 0.42$\pm$0.06 & 23.88   \\ [0.5ex]
    		\hline\hline
	\end{tabular}
	\caption{Classification experiment with \textbf{Cifar-10}. Selective risk (0-1\% error) for various coverage rates.}
	\label{table:cifar10}
\end{table*}

\section{Experimental Design and Details}
\label{sec:implementation}
Our complete code can be downloaded from the following link, \href{https://github.com/geifmany/SelectiveNet}{https://github.com/geifmany/SelectiveNet}.
This section describes the experimental details: datasets used, baseline algorithms used for comparison purposes, and our choice of architectures and hyperparamers. 
\subsection{Datasets}

\textbf{Street View House Numbers (SVHN).} The SVHN dataset \cite{netzer2011reading} is an image classification dataset containing 73,257 training images and 26,032 test images classified into 10 classes representing digits. The images are digits of house street numbers, which are cropped and aligned, taken from the Google Street View service. Image size is $32\times 32\times 3$ pixels.

\textbf{CIFAR-10.} The CIFAR-10 dataset \cite{krizhevsky2009learning} is an image classification dataset comprising a training set of 50,000 images and 10,000 test images classified into 10 categories. The image size is $32\times 32\times 3$ pixels (RGB images). 

\textbf{Cats vs. Dogs.} The Cats vs. Dogs is an image classification dataset extracted from the ASIRRA dataset. It contains 25,000 images of cats and dogs,
12,500 in each class. We randomly split this dataset into a training set containing 20,000 images, and a test set
of 5000 images in a stratified fashion. We also rescaled each image to size 64x64. The average dimension size
of the original images is roughly 360x400.

\textbf{Concrete Compressive Strength.} The Concrete Compressive Strength dataset is a regression dataset from the UCI repository \cite{Dua:2017}. It contains 1030 instances with eight numerical features and one target value. The target value is the compressive strength of concrete, which is predicted based on its ingredients (seven features) and age (one feature).

\subsection{Baseline Methods}
We compare the proposed method with two baselines: SR and MC-dropout, which are described below. 

\textbf{Softmax Response (SR).} The SR method estimates prediction confidence  by the maximum softmax value activation for a given  instance. According to \citet{geifman2017selective}, SR, which is the most straightforward method, is a top performer in selective prediction.

\textbf{Monte Carlo-dropout (MC-Dropout) \cite{gal2016dropout}.} Prediction confidence is quantified using variance statistics of  multiple feed-forward passes of an instance through the network with dropout applied. In both classification and regression, MC-dropout was implemented as recommended by its creators. Specifically,  in classification, we used a dropout rate of $p=0.5$ and 100 feed-forward MC iterations. In regression, we used a dropout rate of $p=0.05$ and 200 feed-forward MC iterations.

\begin{table*}[h!]
	\centering
	\begin{tabular}{c c c c c c} 
		\hline
		\hline
		Coverage&  SelectiveNet risk& MC-dropout risk&  \% improvement &SR risk & \% improvement\\
		\hline\hline
1.00 & {3.21$\pm$0.08} & 3.21$\pm$0.08 & 0.00  & 3.21$\pm$0.08 & 0.00    \\ [0.5ex]
0.95 & {1.40$\pm$0.01} & 1.40$\pm$0.05 & 0.00  & 1.39$\pm$0.05 & -0.77   \\ [0.5ex]
0.90 & \textbf{0.82$\pm$0.01} & 0.90$\pm$0.04 & 9.05  & 0.89$\pm$0.04 & 8.47   \\ [0.5ex]
0.85 & \textbf{0.60$\pm$0.01} & 0.71$\pm$0.03 & 15.26  & 0.70$\pm$0.03 & 13.61   \\ [0.5ex]
0.80 & \textbf{0.53$\pm$0.01} & 0.61$\pm$0.01 & 14.07  & 0.61$\pm$0.02 & 14.07   \\ [0.5ex]
    		\hline\hline
	\end{tabular}
	\caption{Classification experiment with \textbf{SVHN}. Selective risk (0-1\% error) for various coverage rates.}
	\label{table:svhn}
\end{table*}

\subsection{Architectures and Hyperparameters}
For the convolutional neural network (CNN) experiments, we used the well-known VGG-16 architecture \cite{simonyan2014very}, optimized for the small datasets and image sizes as suggested by \citet{liu2015very}. The  changes from the original VGG-16 architecture are: (i) we used  only one fully connected layer with 512 neurons (the original VGG-16 has two fully connected layers of 4096 neurons). (ii) we added batch normalization \cite{ioffe2015batch} (iii) we added dropout \cite{srivastava2014dropout}. We used standard data augmentation consisting of horizontal flips, vertical and horizontal shifts, and rotations. The network was optimized using \emph{stochastic gradient descent} (SGD) with a momentum of 0.9, an initial learning rate of 0.1, and a weight decay of 5e-4. The learning rate was reduced by 0.5 every 25 epochs, and trained for 300 epochs. With this setting, we reached \emph{full coverage} validation accuracy of 96.79\% for SVHN, 93.21\% for Cifar-10 and 96.42\% for Cats vs. Dogs. These are close to the best known results for the VGG-16 architecture.

The main body block of SelectiveNet is the VGG-16 architecture (see Figure~\ref{fig:architecture}). Both the prediction ($f$) and auxiliary heads ($h$) of SelectiveNet  are  fully-connected softmax layers. The selection head ($g$) is a fully connected hidden layer with $512$ neurons, batch normalization and ReLU activation, followed by a fully connected layer to one output neuron with sigmoid activation. The value of $\alpha$ (the convex combination between the selective loss and the auxiliary loss) was set to $0.5$ for all experiments, and $\lambda$ was set to $32$ (this value was found to be  large enough to preserve the constraint through the training process).

In the regression experiment, which considers a tabular dataset from the UCI repository, we used a fully connected network as the main body block. This block consists of one hidden layer of 64  ReLU activated neurons. Both the prediction  and  auxiliary heads are  fully connected layers with one linearly activated neuron. The selection head $g$ was computed with one hidden layer, with 16 ReLU activated neurons,  followed by a fully connected layer with one Sigmoid activated neuron. Batch normalization was applied in every hidden layer on all modules (main body block, prediction, auxiliary and selection heads).  The regression model was optimized  using the ADAM algorithm \cite{kingma2014adam}  with a learning rate of $5 \times 10^{-4}$ and mini-batch size of 256 and 800 epochs. We used squared loss with  weight decay of $1\times 10^{-4}$ during optimization.

\begin{table*}[h!]
	\centering
	\begin{tabular}{c c c c c c} 
		\hline
		\hline
		Coverage&  SelectiveNet risk& MC-dropout risk&  \% improvement &SR risk & \% improvement\\
		\hline\hline
1.00 & {3.58$\pm$0.04} & {3.58$\pm$0.04} & 0.00  & {3.58$\pm$0.04} & 0.00   \\ [0.5ex]
0.95 & \textbf{1.62$\pm$0.05} & 1.92$\pm$0.06 & 15.40  & 1.91$\pm$0.08 & 15.09    \\ [0.5ex]
0.90 & \textbf{0.93$\pm$0.01} &1.10$\pm$0.05 & 16.13 & 1.10$\pm$0.08 & 15.56      \\ [0.5ex]
0.85 & \textbf{ 0.56$\pm$0.02} & 0.78$\pm$0.06 & 28.30& 0.82$\pm$0.06 & 32.40     \\ [0.5ex]
0.80 & \textbf{0.35$\pm$0.09} & 0.55$\pm$0.02 & 36.39 & 0.68$\pm$0.05 & 48.16     \\ [0.5ex]
	\hline\hline
	\end{tabular}
	\caption{Classification experiment with \textbf{Cats vs. Dogs}. Selective risk (0-1 \% error)  for various coverage rates.}
	\label{tab:cats}
\end{table*}

\begin{table*}[htb]
	\centering
	\begin{tabular}{c c c c} 
		\hline
		\hline
		Coverage&  SelectiveNet risk& MC-dropout risk &   \% improvement\\
		\hline\hline
1.00 &  38.45$\pm$0.90 & 39.01$\pm$0.70 & 1.43   \\ [0.5ex]
0.90 &  35.35$\pm$1.31 & 36.97$\pm$0.46 &4.39   \\ [0.5ex]
0.80 & \textbf{30.48$\pm$0.93} &35.53$\pm$0.61 &  14.23   \\ [0.5ex]
0.70 &  \textbf{27.94$\pm$1.12} & 33.70$\pm$0.58 & 17.09   \\ [0.5ex]
0.60 & \textbf{27.12$\pm$0.99} & 31.23$\pm$0.71 & 13.18   \\ [0.5ex]
0.50 & \textbf{26.81$\pm$1.36} &  28.90$\pm$0.77 & 7.23   \\ [0.5ex]
    		\hline\hline
	\end{tabular}
	\caption{Regression experiment with \textbf{Concrete Compressive Strength}. Selective risk (MSE) for various coverage rates.}
	\label{table:regression}
\end{table*}

\begin{table*}[htb]
	\centering

	\begin{tabular}{c | c c c c c} 
	
	&\multicolumn{5}{c}{Calibration}\\
		 Training&   0.70 &0.75& 0.80& 0.85&  0.90\\
		\hline
0.70 &$ \mb{0.32\pm 0.01}$ & $0.55 \pm 0.00$      & ${0.88 \pm 0.05}$   & $1.48 \pm 0.05$ & $2.65 \pm 0.11$ \\ [0.5ex]
0.75 & $\mb{0.31 \pm 0.04}$ & $\mb{0.48 \pm 0.02}$ & $\mb{0.85 \pm 0.04}$  & $1.47 \pm 0.07$ & $2.63 \pm 0.06$ \\ [0.5ex]
0.80 & $\mb{0.34 \pm 0.01}$ &$ 0.55 \pm 0.04$     & $\mb{0.86 \pm 0.06}$   & $\mb{1.42 \pm 0.05}$ & $\mb{2.40 \pm 0.09}$ \\ [0.5ex]
0.85 & $0.37 \pm 0.02 $& $0.54 \pm 0.03$          & $\mb{0.84 \pm 0.04} $  & $\mb{1.43 \pm 0.08}$ &$ \mb{2.47 \pm 0.07}$ \\ [0.5ex]
0.90 & $0.51 \pm 0.12$ & $0.58 \pm 0.09$          & $\mb{0.79 \pm 0.03} $  &$ \mb{1.35 \pm 0.03}$ & $\mb{2.43 \pm 0.08}$ \\
	\end{tabular}
	\caption{Selective risk of SelectiveNet trained on the Cifar-10 dataset where rows represent a training coverage value and columns represent post-training calibration coverage values. All results were multiplied by 100 and correspond to \% of errors.}
	\label{table:confusion}
\end{table*}

\section{Experiments}
\label{sec:experiments}
In this section we report on the results of several experiments that validate the effectiveness of SelectiveNet. We begin with three image classification experiments where the main body block (see Section \ref{sec:selectivenet}) has a CNN architecture. We then experiment with a regression task where the main body block comprises a  fully connected architecture. 
\subsection{Selective Classification}
We consider  the Cifar-10 dataset first. We trained a version of SelectiveNet whose main body block is based on the VGG-16 architecture;
see the full description of our architecture and implementation details in Section~\ref{sec:implementation}. 
We compared calibrated SelectiveNet instances, which were trained
with coverage rates  $c \in \{0.7,0.75,0.8,0.85,0.9,0.95\}$ to two standard selective classifiers over these coverage rates.
The standard classifiers were constructed using the well-known SR and MC-dropout confidence estimates applied to a trained network 
consisting of a multi-class head (identical to $f$ in SelectiveNet) on top of the same main body block of SelectiveNet.
The standard model was trained on Cifar-10 at full coverage as proposed in \cite{geifman2017selective}.
The results are presented in Table~\ref{table:cifar10}. It can be seen that SelectiveNet significantly outperforms the two baseline methods (SR and MC-dropout). The relative advantage of SelectiveNet varies from 8.5\% at 0.95 coverage compared to SR, and a maximal advantage of $26.8\%$ compared to MC-dropout at 0.75 coverage. It is also evident that the MC-dropout and SR performance is quantitatively  similar as was already reported in \cite{geifman2017selective}.

We turn now to the SVHN dataset. Here again, we trained SelectiveNet with the same VGG-16-based architecture for coverage rates $c \in \{0.8,0.85,0.9,0.95\}$ (coverage rates smaller than 0.8 result in nearly perfect models in this dataset). 
The results in Table~\ref{table:svhn} show SelectiveNet's significant superiority for all coverage rates, with the  exception of the $0.95$ coverage, where all methods are statistically indistinguishable. The maximal relative advantage over this dataset is $14.07\%$ for 0.8 coverage rate. We also observe that here again, the performance of SR and MC-dropout is similar for all coverage rates.

Finally, for the Cats vs. Dogs dataset, whose input image size is larger (64 $\times$ 64), we adapted the same architecture (to this image size) for all the 
contenders and repeated the same experiments with the same coverage rates as in 
the SVHN experiment. Presented in Table~\ref{tab:cats}, the results are unsurprising and again indicate a consistent and significant advantage of
SelectiveNet over the two baselines. 

\begin{figure*}[t!]
	\begin{center}
		\subfigure[Cifar-10]{\includegraphics[width=0.45\linewidth]{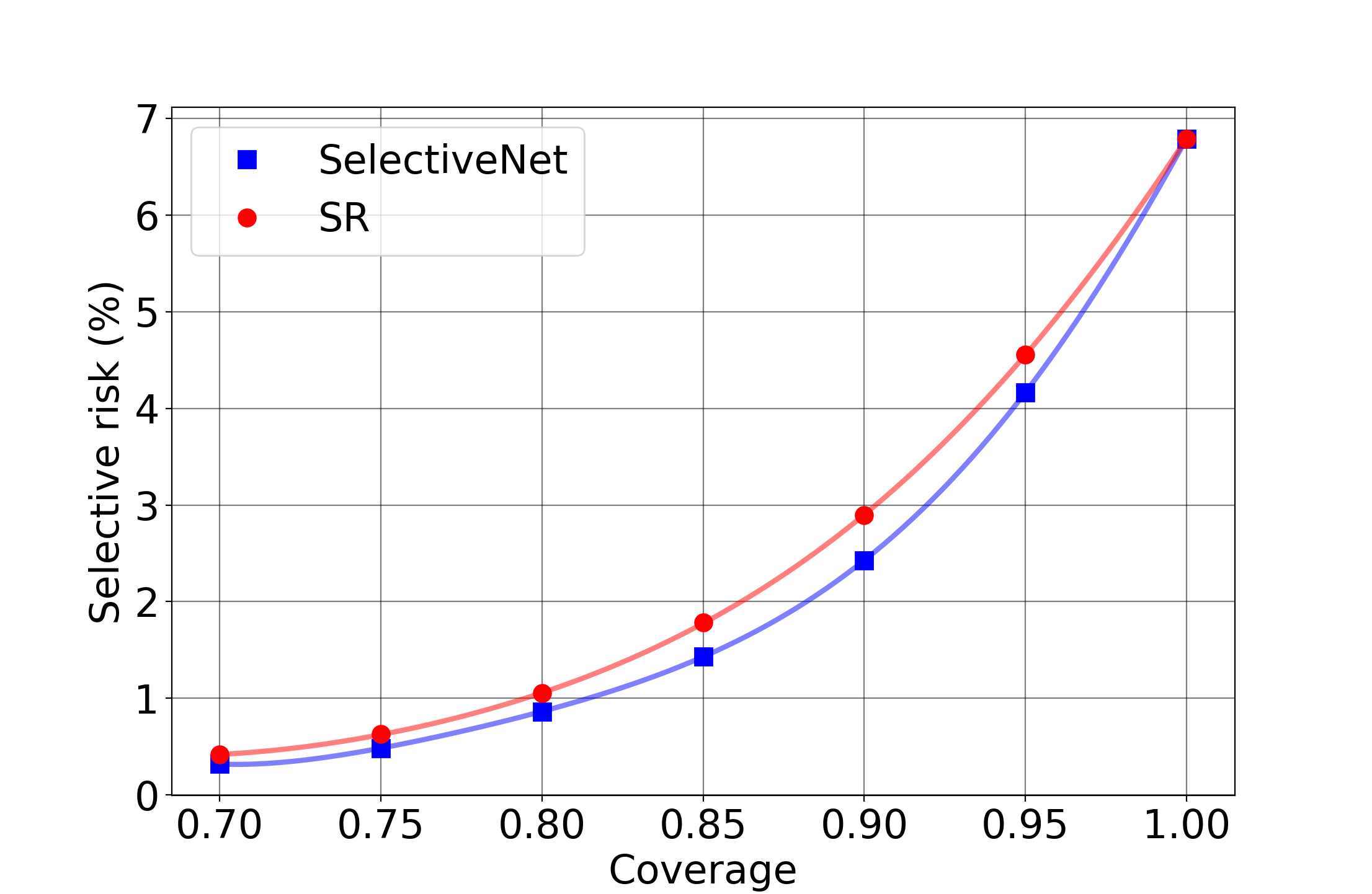}}
		\subfigure[Cats vs. dogs]{\includegraphics[width=0.45\linewidth]{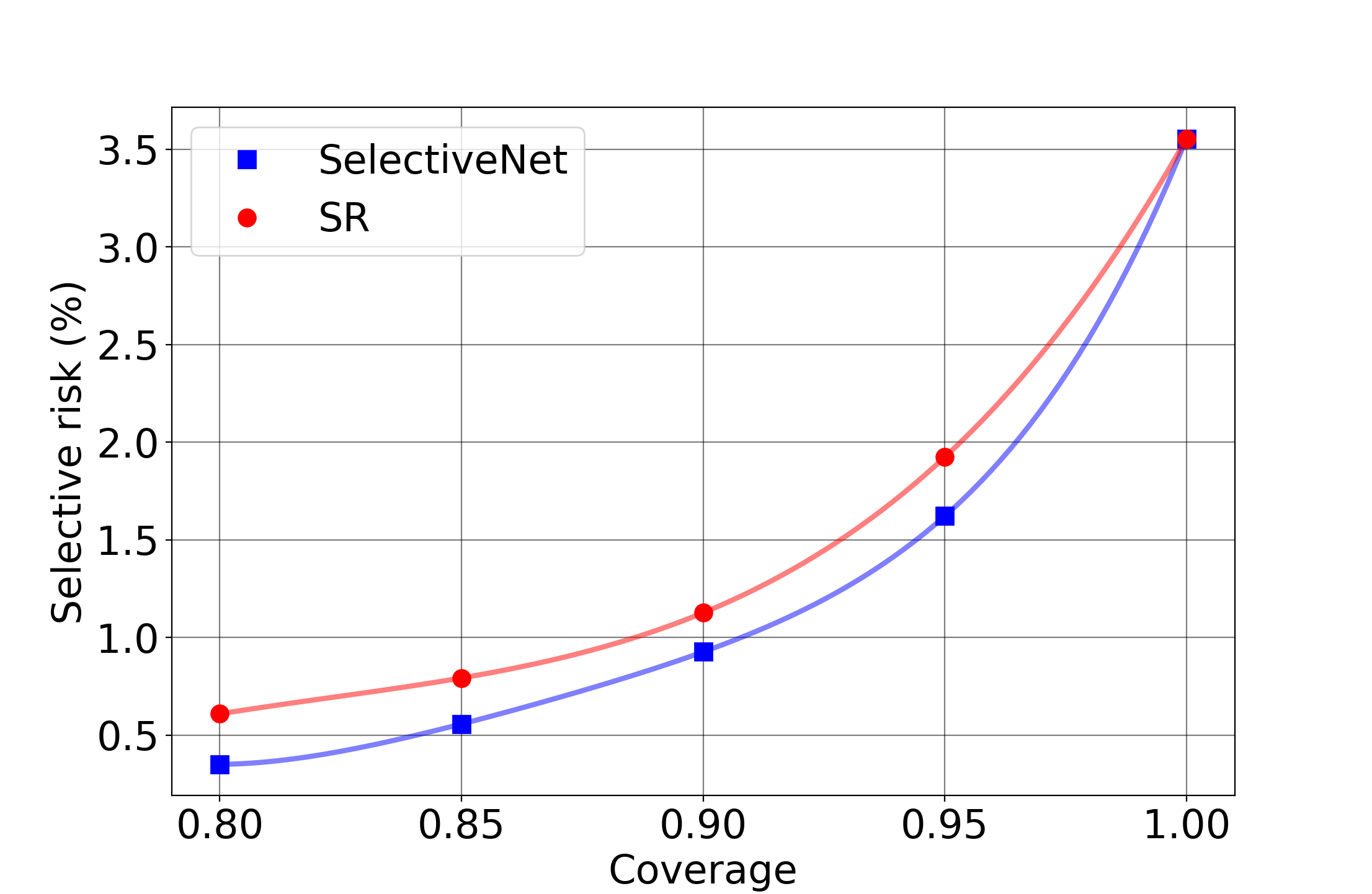} }
	\end{center}
	\caption{Risk coverage curves comparing SelectiveNet and SR over (a) Cifar-10 and (b) Cats vs. Dogs}
	\label{fig:riskCover}
\end{figure*}

Risk-coverage curves depicting the performance of SelectiveNet and SR for  the Cifar-10 and Cats vs. Dogs datasets appear in Figure~\ref{fig:riskCover}. 
These curves nicely demonstrate the consistent advantage of SelectiveNet.

\subsection{Selective Regression}
We ran the regression experiment over the Concrete Compressive Strength dataset from the UCI repository.
For this set we trained a SelectiveNet variant adapted for a regression of tabular data. 
The main body block is now a fully connected neural network and the prediction heads utilize a linear unit as described in Section~\ref{sec:implementation}. The results appear in Table~\ref{table:regression}. 
Clearly, the SR baseline cannot be applied in this setting because there is no softmax layer in regression networks.
We, therefore, compared SelectiveNet only to the MC-dropout method,  currently the only non-ensemble/Bayesian method for deep selective regression.  In this task, SelectiveNet exhibits a distinct advantage over MC-dropout
for all coverage rates below 0.9. There, SelectiveNet  significantly outperforms MC-dropout, with a relative advantage that varies between 26.81\% to 30.48\%.

\begin{figure*}[t!]
	\begin{center}
		\subfigure[SelectiveNet]{\includegraphics[width=0.37\linewidth]{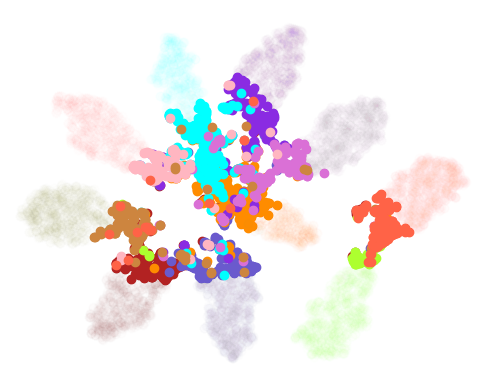}}
		\subfigure[Softmax response]{\includegraphics[width=0.37\linewidth]{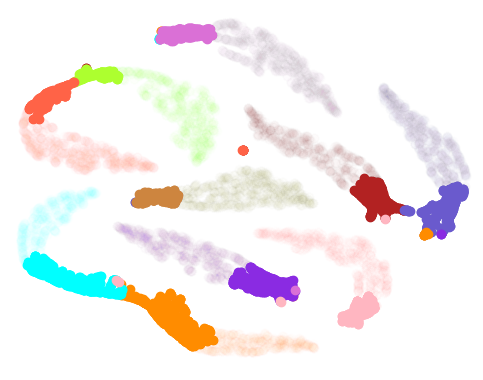} }
		\subfigure[Legend]{\includegraphics[width=0.12\linewidth]{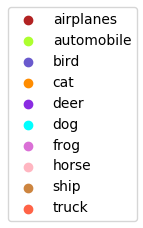}}

	\end{center}
	\caption{t-SNE visualization of the embedding representation of (a) SelectiveNet and (b) SR.}
	\label{fig:tsne}
\end{figure*}
\section{Empirical Observations}
We now empirically analyze some properties of SelectiveNet, related to calibration and  representation structure.
\subsection{Coverage Calibration}
We evaluate the effectiveness and limitations of the post-training coverage calibration technique. We used this calibration to evaluate five models for Cifar-10 trained with five different coverage rates, $c\in \{0.7, 0.75, 0.8, 0.85, 0.9\}$. We calibrated each  model to match each  coverage value. 
Table~\ref{table:confusion} is the resulting coverage ``confusion matrix'' where
element $(i,j)$ in the table is the selective risk of model $i$ calibrated to 
coverage $j$.
For example, the third value in the last column (2.40) is the selective risk of 
SelectiveNet that was optimized with coverage 0.8 during training and  then 
calibrated for coverage 0.9. All the numbers inside the table were multiplied by 100 and reflect the error percent. Each selective risk in the table 
is an average over three independent runs. 

The minimal element in each column appears in bold red, and all other elements in that column, which have one standard error overlap with it, appear in bold red as well.
Thus, the red elements in each column represent all nearly-optimal errors in their 
column. Clearly, the diagonal of this matrix is bold red, indicating that the 
calibrated SelectiveNet, which was trained for coverage $c$, is nearly optimal among 
the other SelectiveNets that were trained with coverage rates other than $c$.
Another interesting observation is that there is no significant compromise due to the 
calibration process, provided that we calibrate towards a coverage rate that is close to the trained (target) coverage.
The fact that the off-diagonal elements often tend to admit inferior selective risks
indicates that the calibrated SelectiveNet effectively optimizes its coverage-specific selective risk.
\subsection{Learned Representation}
It is interesting to inspect  the representation learned by SelectiveNet and those learned by  standard methods such as SR or MC-dropout. Considering SelectiveNet's architecture, one may expect that the main body block will produce  features  that are effective for both the prediction and rejection tasks. In contrast, the standard method relies on standard models that are oblivious to coverage considerations.
To this end, we use the well-known t-Distributed Stochastic Neighbor Embedding (t-SNE) technique \cite{maaten2008visualizing} to visualize the learned representation layer of SelectiveNet and SR that were trained for the Cifar-10 dataset  with $c=0.7$ . We computed the t-SNE transformation of the embedding representation consisting of all 512 neurons in the embedding layer. 
The results are depicted in Figure~\ref{fig:tsne}. In Figure~\ref{fig:tsne}(b) we see the t-SNE of 
the SR model. 
Each color in the figure corresponds to a class label (among the 10 classes in Cifar-10).
For example, the turquoise cluster is `dog' and the orange cluster corresponds to `cat'.
Rejected instances appear as dots in bold colors and the covered points are faded dots.
Similarly, Figure~\ref{fig:tsne}(a) presents the t-SNE of SelectiveNet.

The striking observation is that the SR clusters are relatively well separated regardless 
of rejection but, in contrast, the rejected instances in the SelectiveNet case are very weakly separated.
For example, we see severe confusion between rejected points of classes `airplane' (maroon), 
`ship' (light brown), and `bird' (blue).  Similarly, we see that the rejected instances
of the rest of the animals (`cat',
`dog', `deer', `horse' and `frog') are not well separated as well.
A plausible interpretation of this phenomenon is that SelectiveNet is not wasting 
representational capacity to separate between rejected points, which may leave more capacity 
to learn relevant features in order to separate non-rejected points. 
Another observation is that most of the rejected points were positioned in a central 
cluster. This structure can make it easier for these points to be captured by the selection function $g$.

\section{Concluding Remarks}
We presented SelectiveNet, an effective deep model for selective classification and regression.
The power of this new model stems from the mutual training of both the selection and prediction 
models within the same network, thus forcing the models to focus on the more relevant instances
that will not be rejected in production. Known techniques, such as MC-dropout or SR,
which typically compose a rejection mechanism over a standard prediction model (trained with full coverage), cannot benefit from this advantage.
Our empirical results indicate that the new model currently provides 
the most accurate classification for a given abstention rate.
For selective \emph{regression}, SelectiveNet provides the best and only fast solution, where  
the only other (inferior) alternative is MC-dropout. 
This result motivates the use of SelectiveNet in various deep regression 
applications that require fast inference, such as visual tracking or detection.

We leave a number of issues for future research. As is the case in many  deep learning applications,
also in selective prediction, ensembling techniques improve performance of basic models.
In \citet{lakshminarayanan2017simple}, it was recently shown that an ensemble of 2-15 models can improve upon the basic SR rejection method. It would be interesting to check
if ensembling can improve SelectiveNet as well.
Another interesting direction is to examine the influence of  capacity (architecture choice) of 
the selection function $g$ on the overall performance of SelectiveNet. Here, we have not tried to optimize this choice, but we anticipate that different datasets and perhaps different coverage rates
may require different architectures also for the selection head.
Finally,  considering the known relationship between active learning and selective classification
\cite{el2012active,gelbhart2017relationship},
an important open question is whether the extra power
of SelectiveNet 
can be used to leverage active learning
\cite{geifman2017deep,geifman2018deep}.

\section*{Acknowledgments}
This research was supported by The Israel Science Foundation (grant No. 710/18).

\bibliography{selectivenet_bib}

\begin{thebibliography}{29}
\providecommand{\natexlab}[1]{#1}
\providecommand{\url}[1]{\texttt{#1}}
\expandafter\ifx\csname urlstyle\endcsname\relax
  \providecommand{\doi}[1]{doi: #1}\else
  \providecommand{\doi}{doi: \begingroup \urlstyle{rm}\Url}\fi

\bibitem[Chow(1957)]{chow1957optimum}
Chow, C.~K.
\newblock An optimum character recognition system using decision functions.
\newblock \emph{IRE Transactions on Electronic Computers}, \penalty0
  (4):\penalty0 247--254, 1957.

\bibitem[Cordella et~al.(1995)Cordella, De~Stefano, Tortorella, and
  Vento]{cordella1995method}
Cordella, L.~P., De~Stefano, C., Tortorella, F., and Vento, M.
\newblock A method for improving classification reliability of multilayer
  perceptrons.
\newblock \emph{IEEE Transactions on Neural Networks}, 6\penalty0 (5):\penalty0
  1140--1147, 1995.

\bibitem[Cortes et~al.(2016{\natexlab{a}})Cortes, DeSalvo, and
  Mohri]{cortes2016boosting}
Cortes, C., DeSalvo, G., and Mohri, M.
\newblock Boosting with abstention.
\newblock In \emph{Advances in Neural Information Processing Systems}, pp.\
  1660--1668, 2016{\natexlab{a}}.

\bibitem[Cortes et~al.(2016{\natexlab{b}})Cortes, DeSalvo, and
  Mohri]{cortes2016learning}
Cortes, C., DeSalvo, G., and Mohri, M.
\newblock Learning with rejection.
\newblock In \emph{International Conference on Algorithmic Learning Theory},
  pp.\  67--82. Springer, 2016{\natexlab{b}}.

\bibitem[De~Stefano et~al.(2000)De~Stefano, Sansone, and Vento]{de2000reject}
De~Stefano, C., Sansone, C., and Vento, M.
\newblock To reject or not to reject: that is the question-an answer in case of
  neural classifiers.
\newblock \emph{IEEE Transactions on Systems, Man, and Cybernetics, Part C
  (Applications and Reviews)}, 30\penalty0 (1):\penalty0 84--94, 2000.

\bibitem[Dheeru \& Karra~Taniskidou(2017)Dheeru and Karra~Taniskidou]{Dua:2017}
Dheeru, D. and Karra~Taniskidou, E.
\newblock {UCI} machine learning repository, 2017.
\newblock URL \url{http://archive.ics.uci.edu/ml}.

\bibitem[El-Yaniv \& Wiener(2010)El-Yaniv and Wiener]{ElYaniv10}
El-Yaniv, R. and Wiener, Y.
\newblock On the foundations of noise-free selective classification.
\newblock \emph{Journal of Machine Learning Research}, 11:\penalty0 1605--1641,
  2010.

\bibitem[El-Yaniv \& Wiener(2012)El-Yaniv and Wiener]{el2012active}
El-Yaniv, R. and Wiener, Y.
\newblock Active learning via perfect selective classification.
\newblock \emph{Journal of Machine Learning Research (JMLR)}, 13\penalty0
  (Feb):\penalty0 255--279, 2012.

\bibitem[El-Yaniv \& Wiener(2015)El-Yaniv and Wiener]{wiener2015agnostic}
El-Yaniv, R. and Wiener, Y.
\newblock Agnostic pointwise-competitive selective classification.
\newblock \emph{Journal of Artificial Intelligence Research}, 52:\penalty0
  171--201, 2015.

\bibitem[Fumera \& Roli(2002)Fumera and Roli]{fumera2002support}
Fumera, G. and Roli, F.
\newblock Support vector machines with embedded reject option.
\newblock In \emph{Pattern recognition with support vector machines}, pp.\
  68--82. Springer, 2002.

\bibitem[Gal \& Ghahramani(2016)Gal and Ghahramani]{gal2016dropout}
Gal, Y. and Ghahramani, Z.
\newblock Dropout as a bayesian approximation: representing model uncertainty
  in deep learning.
\newblock In \emph{Proceedings of The 33rd International Conference on Machine
  Learning}, pp.\  1050--1059, 2016.

\bibitem[Geifman \& El-Yaniv(2017{\natexlab{a}})Geifman and
  El-Yaniv]{geifman2017deep}
Geifman, Y. and El-Yaniv, R.
\newblock Deep active learning over the long tail.
\newblock \emph{arXiv preprint arXiv:1711.00941}, 2017{\natexlab{a}}.

\bibitem[Geifman \& El-Yaniv(2017{\natexlab{b}})Geifman and
  El-Yaniv]{geifman2017selective}
Geifman, Y. and El-Yaniv, R.
\newblock Selective classification for deep neural networks.
\newblock In \emph{Advances in neural information processing systems}, pp.\
  4878--4887, 2017{\natexlab{b}}.

\bibitem[Geifman \& El-Yaniv(2018)Geifman and El-Yaniv]{geifman2018deep}
Geifman, Y. and El-Yaniv, R.
\newblock Deep active learning with a neural architecture search.
\newblock \emph{arXiv preprint arXiv:1811.07579}, 2018.

\bibitem[Geifman et~al.(2018)Geifman, Uziel, and El-Yaniv]{geifman2018bias}
Geifman, Y., Uziel, G., and El-Yaniv, R.
\newblock Bias-reduced uncertainty estimation for deep neural classifiers.
\newblock 2018.

\bibitem[Gelbhart \& El-Yaniv()Gelbhart and El-Yaniv]{gelbhart2017relationship}
Gelbhart, R. and El-Yaniv, R.
\newblock The relationship between agnostic selective classification active
  learning and the disagreement coefficient.
\newblock \emph{JMLR}, \penalty0 (20(33):1−38).

\bibitem[Hellman(1970)]{hellman1970nearest}
Hellman, M.~E.
\newblock The nearest neighbor classification rule with a reject option.
\newblock \emph{IEEE Transactions on Systems Science and Cybernetics},
  6\penalty0 (3):\penalty0 179--185, 1970.

\bibitem[Hoeffding(1963)]{Hoeffding:1963}
Hoeffding, W.
\newblock Probability inequalities for sums of bounded random variables.
\newblock \emph{Journal of the American Statistical Association}, 58\penalty0
  (301):\penalty0 13--30, March 1963.
\newblock URL \url{http://www.jstor.org/stable/2282952?}

\bibitem[Ioffe \& Szegedy(2015)Ioffe and Szegedy]{ioffe2015batch}
Ioffe, S. and Szegedy, C.
\newblock Batch normalization: Accelerating deep network training by reducing
  internal covariate shift.
\newblock \emph{arXiv preprint arXiv:1502.03167}, 2015.

\bibitem[Kingma \& Ba(2014)Kingma and Ba]{kingma2014adam}
Kingma, D.~P. and Ba, J.
\newblock Adam: A method for stochastic optimization.
\newblock \emph{arXiv preprint arXiv:1412.6980}, 2014.

\bibitem[Krizhevsky \& Hinton(2009)Krizhevsky and
  Hinton]{krizhevsky2009learning}
Krizhevsky, A. and Hinton, G.
\newblock Learning multiple layers of features from tiny images.
\newblock 2009.

\bibitem[Lakshminarayanan et~al.(2017)Lakshminarayanan, Pritzel, and
  Blundell]{lakshminarayanan2017simple}
Lakshminarayanan, B., Pritzel, A., and Blundell, C.
\newblock Simple and scalable predictive uncertainty estimation using deep
  ensembles.
\newblock In \emph{Advances in Neural Information Processing Systems}, pp.\
  6402--6413, 2017.

\bibitem[Liu \& Deng(2015)Liu and Deng]{liu2015very}
Liu, S. and Deng, W.
\newblock Very deep convolutional neural network based image classification
  using small training sample size.
\newblock In \emph{Pattern Recognition (ACPR), 2015 3rd IAPR Asian Conference
  on}, pp.\  730--734. IEEE, 2015.

\bibitem[Maaten \& Hinton(2008)Maaten and Hinton]{maaten2008visualizing}
Maaten, L. v.~d. and Hinton, G.
\newblock Visualizing data using t-sne.
\newblock \emph{Journal of machine learning research}, 9\penalty0
  (Nov):\penalty0 2579--2605, 2008.

\bibitem[Netzer et~al.(2011)Netzer, Wang, Coates, Bissacco, Wu, and
  Ng]{netzer2011reading}
Netzer, Y., Wang, T., Coates, A., Bissacco, A., Wu, B., and Ng, A.~Y.
\newblock Reading digits in natural images with unsupervised feature learning.
\newblock In \emph{NIPS workshop on deep learning and unsupervised feature
  learning}, volume 2011, pp.\ ~5, 2011.

\bibitem[Potra \& Wright(2000)Potra and Wright]{potra2000interior}
Potra, F.~A. and Wright, S.~J.
\newblock Interior-point methods.
\newblock \emph{Journal of Computational and Applied Mathematics}, 124\penalty0
  (1-2):\penalty0 281--302, 2000.

\bibitem[Simonyan \& Zisserman(2014)Simonyan and Zisserman]{simonyan2014very}
Simonyan, K. and Zisserman, A.
\newblock Very deep convolutional networks for large-scale image recognition.
\newblock \emph{arXiv preprint arXiv:1409.1556}, 2014.

\bibitem[Srivastava et~al.(2014)Srivastava, Hinton, Krizhevsky, Sutskever, and
  Salakhutdinov]{srivastava2014dropout}
Srivastava, N., Hinton, G., Krizhevsky, A., Sutskever, I., and Salakhutdinov,
  R.
\newblock Dropout: a simple way to prevent neural networks from overfitting.
\newblock \emph{The Journal of Machine Learning Research}, 15\penalty0
  (1):\penalty0 1929--1958, 2014.

\bibitem[Wiener \& El-Yaniv(2012)Wiener and El-Yaniv]{wiener2012pointwise}
Wiener, Y. and El-Yaniv, R.
\newblock Pointwise tracking the optimal regression function.
\newblock In \emph{Advances in Neural Information Processing Systems (NIPS)},
  pp.\  2042--2050, 2012.

\end{thebibliography}
\bibliographystyle{icml2019}



\end{document}